\documentclass[10pt,twocolumn,letterpaper]{article}

\usepackage{cvpr}
\usepackage{times}
\usepackage{epsfig}
\usepackage{graphicx}
\usepackage{amsmath}
\usepackage{amssymb}

\usepackage{slashbox}
\usepackage[utf8]{inputenc}

\usepackage{setspace}
\usepackage{algpseudocode}
\usepackage{algorithm}
\usepackage{algorithmicx}
\usepackage{mathtools}
\usepackage{boxedminipage}
\usepackage{graphicx}
\usepackage{verbatim}
\usepackage{times} 
\usepackage{dsfont}
\usepackage{multirow}
\usepackage{amsfonts}
\usepackage[T1]{fontenc}
\usepackage{url}
\usepackage{epsfig}
\usepackage{epstopdf}
\usepackage{caption}
\usepackage{subcaption}
\usepackage{adjustbox}
\usepackage{booktabs}

\usepackage[pagebackref=true,breaklinks=true,letterpaper=true,colorlinks,bookmarks=false]{hyperref}

\cvprfinalcopy 

\def\cvprPaperID{****} 

\ifcvprfinal\fi
\begin{document}

\title{Deep Feature Tracker: A Novel Application for Deep Convolutional Neural Networks}

\author{Mostafa Parchami\\
Ford Motor Company\\
Ann Arbor, Michigan\\
{\tt\small mparchami@ford.com}
\and
Saif~Iftekar~Sayed\\
University of Texas at Arlington\\
Arlington, Texas\\
{\tt\small saififtekar.sayed@mavs.uta.edu}
}

\maketitle

\begin{abstract}

Feature tracking is the building block of many applications such as visual odometry, augmented reality, and target tracking. Unfortunately, the state-of-the-art vision based tracking algorithms fail in surgical images due to the challenges imposed by the nature of such environments. In this paper, we proposed a novel and unified deep learning based approach that can learn how to track features reliably as well as learn how to detect such reliable features for the tracking purpose. The proposed network dubbed as Deep-PT, consists of a  tracker network which is a convolutional neural network simulating cross correlation in terms of deep learning and two fully connected networks that operate on the output of intermediate layers of the tracker to detect features and predict track-ability of the detected points. The ability to detect features based on the capabilities of the tracker distinguishes the proposed method from previous algorithms used in this area and improves the robustness of the algorithms against dynamics of the scene. The network is trained using multiple datasets due to the lack of specialized dataset for feature tracking datasets and extensive comparisons are conducted to compare the accuracy of Deep-PT against recent pixel tracking algorithms. As the experiments suggest, the proposed deep architecture deliberately learns what to track and how to track and outperforms the state-of-the-art methods.

\end{abstract}

\section{Introduction}
Thanks to recent technological advances in robotic assisted surgery especially in minimally-invasive surgery (MIS), endoscopic cameras are nowadays widely used as a tool for diagnosis and cancer treatment procedures. During the MIS, the surgical instruments and the endoscope are inserted through tiny incisions and the surgery is performed remotely from a control console by utilizing video guidance provided by the endoscopic camera. Video-guided surgery has increased the need for translating the traditional computer vision algorithms for surgical vision environment and adapt them with unforeseen challenges available in such environments.

Compared to the traditional open-cavity surgery, in MIS the patients benefit from smaller incisions, less trauma, shorter hospitalization, less pain and more importantly lower infection risks~\cite{lin2015video}. Unfortunately, MIS poses major challenges for the surgeon who will experience a reduced awareness of the patient’s anatomy due to narrow field of view of the endoscopic camera and lost depth perception~\cite{wieringa2014improved}. As a consequence, the surgeon faces difficulty in locating and tracking critical anatomical structures such as blood vessels resulting in a higher risk of accidentally damaging an organ. 

In this regard, computer-assisted navigation systems have been developed during the past years that promise to enhance the surgeon’s perception of the environment by fusing the available pre-operative radiological data with the live endoscopic video. Detecting and tracking visual features in real-time is at the core of any such system to provide guidance and on-line decision-making assistance. Visual feature tracking finds a wide range of applications from target tracking~\cite{pullens2016real},~\cite{liu2013optical},~\cite{amber2015feature} to tool tracking segmentation~\cite{bell2013image},~\cite{garcia2016real}, augmented reality~\cite{marques2015framework}, and deformation recovery~\cite{lin2015video}. 

\begin{figure}[h]
\begin{center}
   \includegraphics[width=0.9\linewidth]{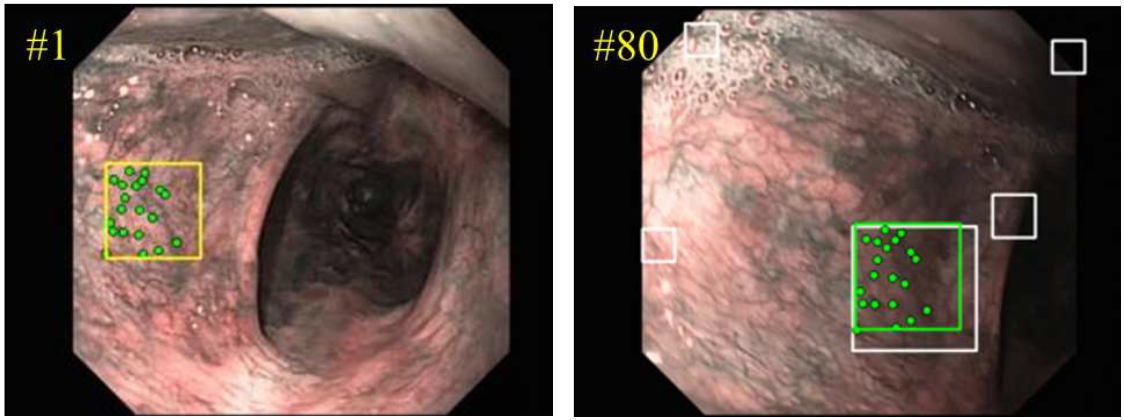}
\end{center}
   \caption{Application of pixel based target tracking in biopsy. Left: The image where an optical biopsy site is selected. Right: The image where the site is tracked from previous frames using tracked keypoints.~\cite{ye2016online}}
\label{fig:TargetTrack}
\end{figure}

For example, Figure~\ref{fig:TargetTrack} shows a target tracking system where tracking of the area of interest is carried out by performing feature tracking on the surface of the organ. Tracking systems usually rely on an external feature detection system that detects a set of good features for tracking purpose. 

\begin{figure}[h]
\begin{center}
   \includegraphics[width=0.9\linewidth]{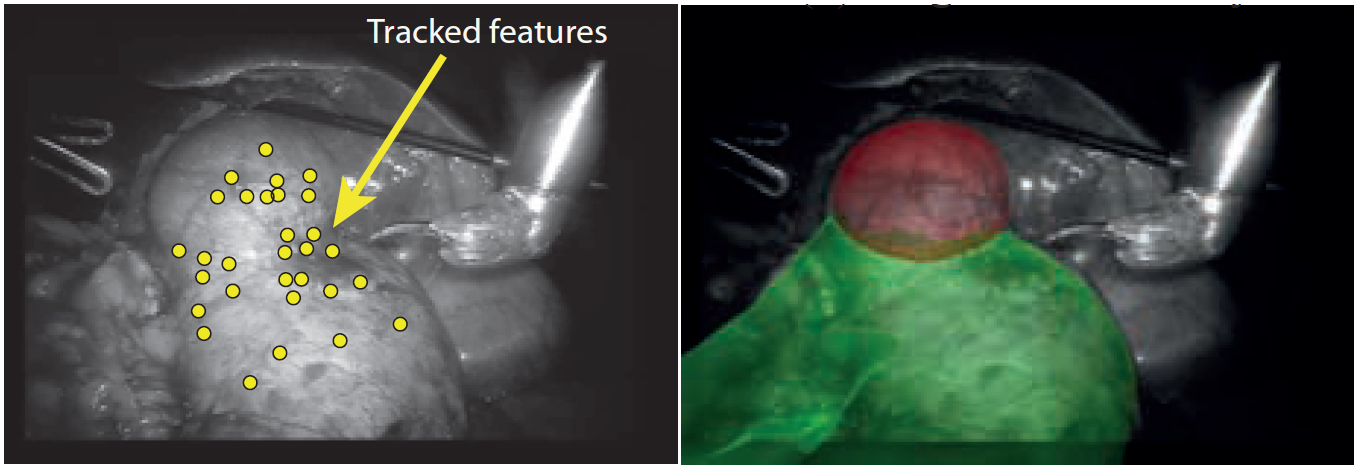}
\end{center}
   \caption{Application of pixel based feature tracking in AR where the tracked pixels are used as anchor points to overlay a pre-operative 3-D model. Left: The tracked points visualized on the current frame. Right: The overlaid CT-scan model on top of the image.~\cite{puerto2012hierarchical}}
\label{fig:AugReal}
\end{figure}

As another example, Figure~\ref{fig:AugReal} shows a scenario where these tracked features can be used as anchor points for overlaying augmented reality on top of the image to give the surgeon a hint of depth perception. In this scenario the pre-operative radiological 3-D model is overlaid on top of the organ and the rendering is then updated by tracking the anchor points over time and aligning the 3-D model accordingly.

Despite recent efforts in adapted well-known feature detecting and tracking algorithms such as Kanade-Lucas-Tomasi (KLT) Tracker, most of the proposed prototypes~\cite{davison2003real},~\cite{mountney2008soft},~\cite{lim2005direct}~\cite{richa2008efficient},~\cite{figl2010image} fail to provide a reliable and accurate long-term tracking under a surgical environment ~\cite{lin2015video}. This is mostly due to the challenges posed by endoscopic imagery such as dynamic nature of the surgical environment, occlusions, sudden tissue deformations, specular highlights, image clutter caused by blood or smoke, and large texture-less areas~\cite{parchami2014endoscopic}. As a result, off-the-shelf computer vision approaches simply fail when applied to the endoscopic images and usually require major revisions in order to make them applicable to such scenarios. Different approaches taken by scientist in order to address poor performance of of KLT includes exploiting Extended Kalman Filter(EKF) to utilize temporal information~\cite{figl2010image}, on-line appearance learning and treating tracking as a classification problem~\cite{mountney2008soft}, Thin Plate Spline (TPS) to track deforming surface~\cite{lim2005direct}, fusing intensity from stereo pair images for intensity matching~\cite{stoyanov2005soft}, hierarchical feature matching~\cite{souza2011adaptive}. 

Each of the aforementioned methods try to improve the accuracy of tracking by tackling the problem from a different perspective. However, the ultimate tracking system should be a self-contained framework that is able to overcome all shortcomings of the state-of-the-art methods. The goal of this thesis is to advance the reliability and robustness of surgical vision methods for endoscopic images by developing real-time algorithms to accurately detect and track reliable features under challenging and dynamic surgical environment.

\section{Proposed Method}
\begin{figure}[h!]
\begin{center}
   \includegraphics[width=0.9\linewidth]{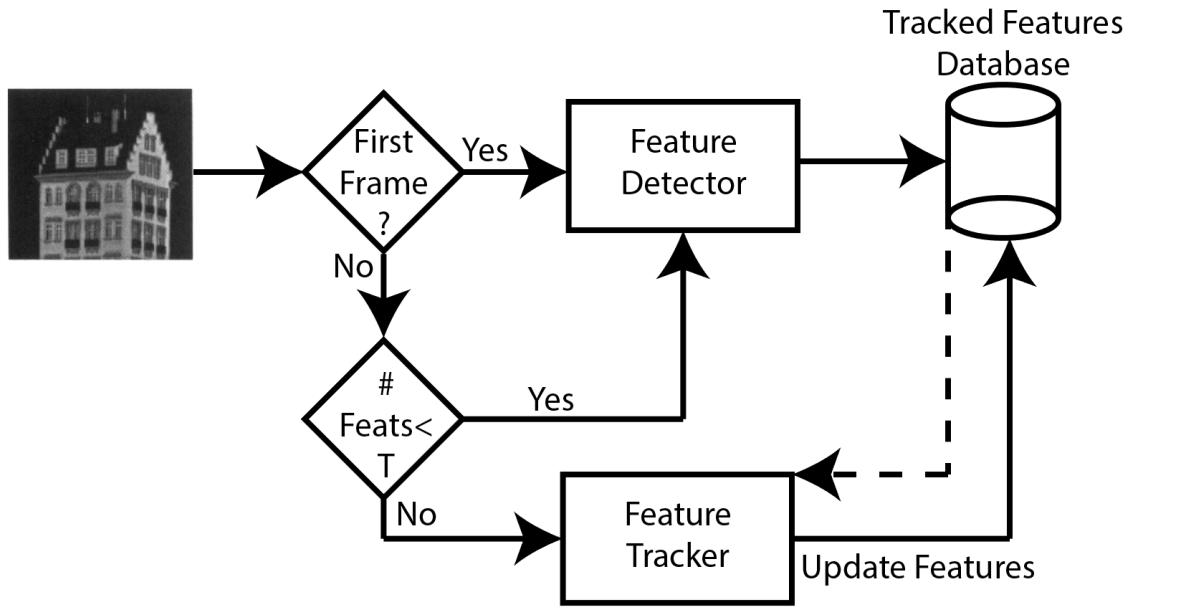}
\end{center}
   \caption{Overall diagram for the Deep-PT. The method takes in input the live video from the single camera and detects and tracks features over time.}
\label{fig:Overall}
\end{figure}

The main diagram for the invention is illustrated in Figure~\ref{fig:Overall} and described in what follows. The framework has two major components. The first component, “Feature Detector” is responsible for detecting trackable features in the image. By trackable, we mean a feature than can be detected and recognized under small motion of camera and changes in the scene such as illumination. The second component, “Feature Tracker”, takes the detected features and localized them in the next frame. In the first frame of the video sequence, the detector finds good features to track (initialization). During the tracking, if the number of the tracked features falls below a threshold ($\epsilon$), then the “Feature Detector” is revoked to detect more features and add them to the list of tracked features (re-initialization).

\subsection{Feature Detection Network}
\begin{figure}[h!]
\begin{center}
   \includegraphics[width=0.9\linewidth]{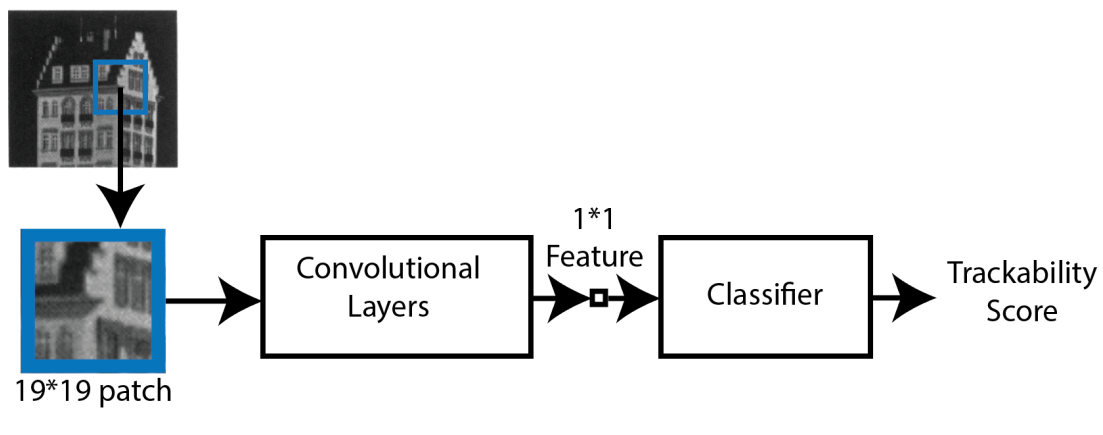}
\end{center}
   \caption{Main diagram for the feature detection pipeline.}
\label{fig:FeatDetect}
\end{figure}

The main diagram for the Feature Detector module is illustrated in Figure~\ref{fig:FeatDetect} and described below in detail. 

The “Feature Detector” module uses a deep convolutional neural network to predict how good the given pixel is for the tracking purpose. It takes a patch around a pixel as input and spits out a trackability score. The 9 convolutional layers extract feature from the patch and a fully connected layer along with a softmax layer calculate a score for the given patch. This network sweeps through all the pixels of the image and evaluate each pixel location for tracking. If the score is higher than a threshold, the location of the pixel (known as feature or or keypoint or interest point) is added to the database of features. One of the advantages of such feature detector is it's low computational burden as the convolutional layers are already applied to the image for tracking purpose. Moreover, if the feature detector is trained based on the capabilities of the tracker, then such unified tracking system can achieve higher accuracy and reliability.

\subsection{Feature Tracking Network}
\begin{figure}[h!]
\begin{center}
   \includegraphics[width=0.9\linewidth]{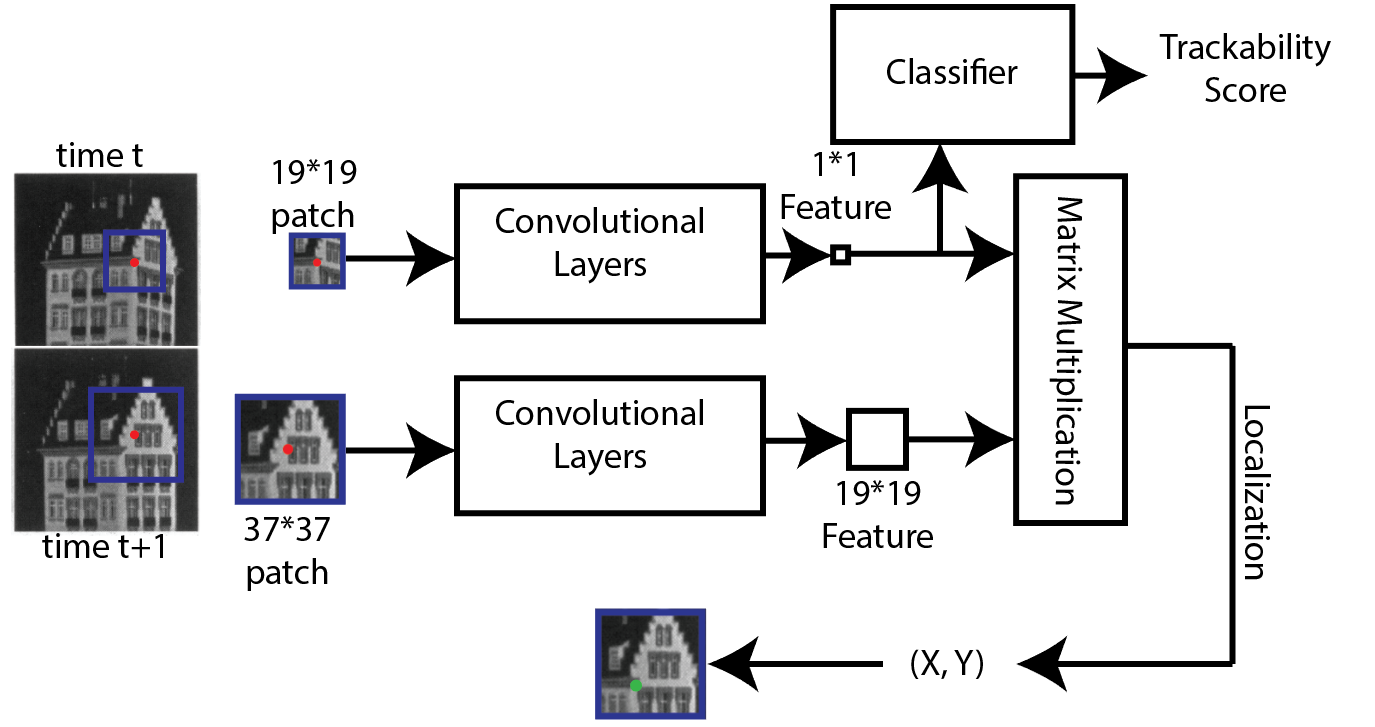}
\end{center}
   \caption{Main diagram for the feature tracking pipeline.}
\label{fig:FeatTrack}
\end{figure}

The main diagram for the Feature Detector module is illustrated in Figure~\ref{fig:FeatTrack} and described below in detail.

The “Feature Tracker” tracks each feature separately and does not consider any spatial correlation. It takes a small patch centered at the given location in the previous frame. Then, the same convolutional layers than the feature detector are applied to extract a representation for the patch. Also, a bigger patch (here 37*37) centered at the same position on the current frame will be passed through the same set of convolutional layers to extract the features. Once the deep representation of the patches are obtained, a matrix multiplication will join these two branches of the network and the location of the maximum in the resulted matrix determines the position of the feature in the current frame. The matrix multiplication resembles the traditional cross-correlation in patch-based matching. On the other hand, a similar fully connected network than the one in the feature detector is applied on the vectorized final matrix to determine tracking score of the feature. During the tracking, if this score is below a threshold for a specific feature, the same feature will not be tracked anymore. This may be caused by distortion, big change in viewpoint, or sudden deformation of the scene. Tracking score functionality allows the tracking framework to adapt itself with the dynamics of the scene and re-initialization gives it reliability to track more features once tracking is considered to be lost.

\subsection{Training The Architecture}
Training the proposed deep architecture requires a large dataset specific to feature tracking. Unfortunately, the lack of such training dataset that is specific to tracking made training even more difficult. Moreover, the network consists of multiple components that should be trained separately on a suitable dataset for each task. Therefore, the training is implemented in three stages: 1) training the tracker, 2) training the tracking score network, 3) training the feature detection network.

\subsubsection{Training The Tracking Network}
In order to train the tracking network, the tracking score network which is a fully connected is detached and the tracker is trained separately. For this purpose, We adopted the KITTI Flow 2012 dataset~\cite{geiger2012we} which contains 389 pair of stereo images with ground truth suitable for stereo reconstruction and visual SLAM. The ground truth data provided by the dataset can be used to generate pairs of corresponding points for each pair of consecutive images. 

In order to avoid training the network with texture-less areas, the training data is generated around Harris corners or SIFT keypoints with a small radius. This will ensure that the training data does not contain any texture-less point such as sky or the road which may bias the tracker. Moreover, points that move more than 19 pixels from the previous frame are dismissed since those points don't satisfy our assumption that pixels don't move more than 19 pixels from a from the previous frame.

\begin{figure*}[th!]
\begin{center}
   \includegraphics[width=0.95\linewidth]{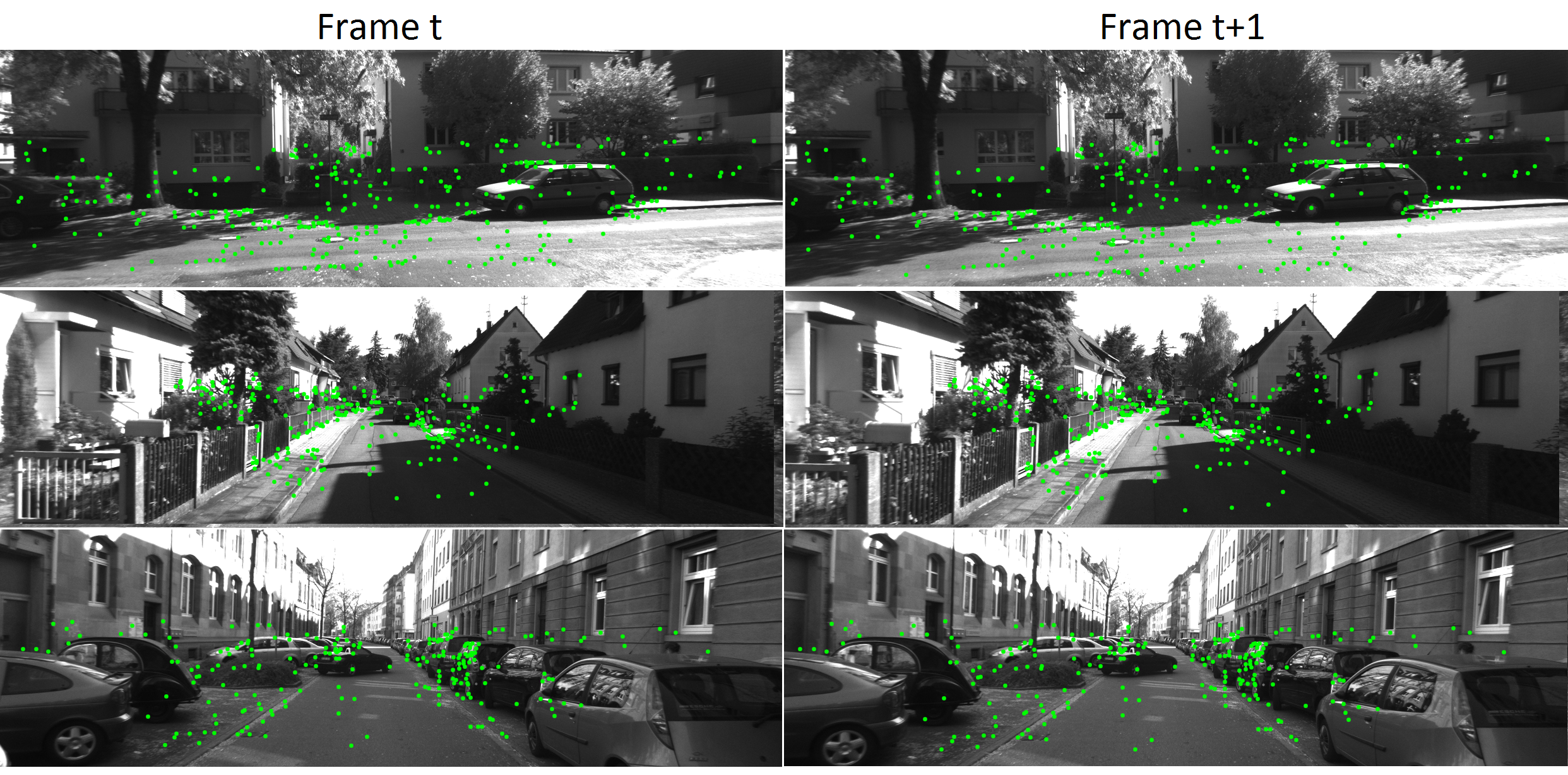}
\end{center}
   \caption{Sample of training points generated for KITTTI Flow 2012 dataset. Each row presents a single pair of consecutive images with features marked with green dots.}
\label{fig:KITTITrain}
\end{figure*}

The architecture is trained using an ad-hoc criterion where a 2-D Gaussian distribution with $\sigma=3 pixels$ centered at the target position in the $frame t+1$ to determine the loss. A small patch is extracted from the $frame t$ where the size of the patch is equal to the network's left branch receptive field. On the other hand, a bigger patch is extracted from the $frame t+1$ where the patch extends to network's receptive size plus small window size of $37 pixels$ for the tracking and this patch is centered at the position of pixel in $frame t$. The inner-product layer of the network produces a score for each location in the patch taken from $frame t+1$ and this allows us to compute a softmax for each pixel over all possible locations in that window. The parameters of the network are updated by minimizing cross entropy-loss with respect to the parameter set $W$ give by:

$min_{W}\sum_{i}\sum_{j} P_{gt}(x_{i},y_{i}) log P_{i}(x_{i},y_{i},W)$

Where $P_{gt}(x_{i},y_{i})$ is a $3\times3$ Gaussian filter centered around the ground truth and zero every where else to consider 3-pixel error metric. Also, $P_{i}(x_{i},y_{i},W)$ is the softmax probability distribution obtained by the forward pass using parameters $W$ at position $(i,j)$ in the window.

Roughly 100K points are used for training the tracker over the coarse of 200 epochs. Once the tracker is trained, it can be used to localize a feature in the next frame if the pixel moves in the $37\times37$ region. Table~\ref{table:TrainingParamsTrack} tabulates the main training parameters used to train the tracker network with Adam algorithm.

Fig.~\ref{fig:KITTITrain} visualizes several images from the KITTI Flow 2012 dataset along with the generated ground truth points. The first column is a cropped region of the original image in the dataset with the location of each keypoint. The second column visualizes the next cropped frame with the same corresponding keypoints in the current frame.

\begin{table}[]
\caption{Tracker's training parameters. Note that in addition to the learning rate decay, the learning rate is decreased by factor of 0.2 every 30 epochs after the epoch number 120.}
\label{table:TrainingParamsTrack}
\begin{tabular}{@{}c|cccc@{}}
\toprule
Parameter                   & Learning             & Learning       & Weight              & Momentum                 \\
 & Rate & Rate Decay & Decay & \\
\midrule
\multicolumn{1}{c|}{Values} & \multicolumn{1}{c|}{1e-2} & \multicolumn{1}{c|}{1e-7} & \multicolumn{1}{c|}{1e-4} & \multicolumn{1}{c|}{0.9} \\ \bottomrule
\end{tabular}
\end{table}
\subsubsection{Training The Tracking Score Network}

\begin{figure}[th!]
\begin{center}
   \includegraphics[width=0.9\linewidth]{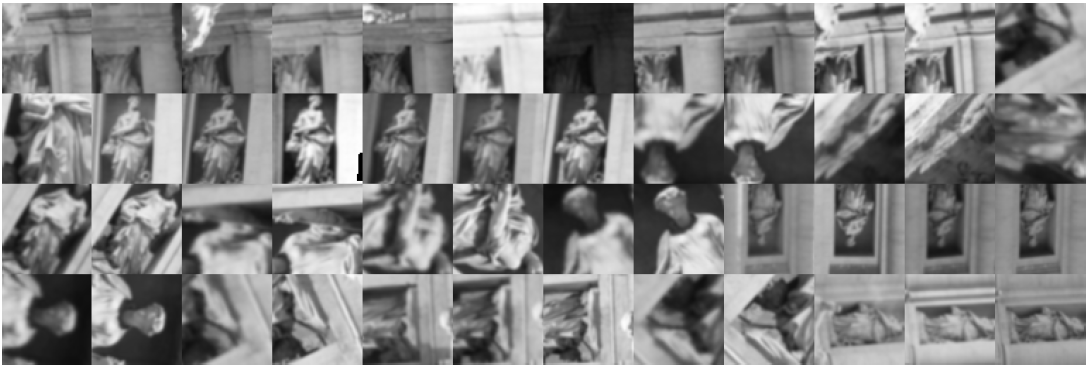}
\end{center}
   \caption{UBC Patches dataset~\cite{han2015matchnet} contains several viewpoints of each 3D point and is challenging due to different levels of rotation, translation and scale.}
\label{fig:UBCPatches}
\end{figure}

Most applications that rely on tracking pixels, require a tracking score to detect when the tracking is lost or a specific feature is not reliable. In order to obtain such information from the tracker, a fully connected network is attached to the output of the matrix dot product layer that will generate a matching score for the two patches. In order to train this network, we adopted the UBC patch dataset~\cite{UBCPatches} which is originally collected for local descriptor learning~\cite{winder2007learning} by Winder et al. Fig.~\ref{fig:UBCPatches} visualizes some challenging images from this dataset where each patch is followed by several patches that represent a single 3-D point captured from different viewpoints. 

In order to be able to compare the network with state-of-the-art methods, the training and testing protocols suggested by~\cite{han2015matchnet} are applied. It's worth mentioning that the parameters of the convolutional layers are not updated during training the tracking score network to make sure the accuracy of the tracker is not deteriorated. Table~\ref{table:TrainingParamsScore} tabulates the main training parameters used to train the tracking score network with Adam algorithm. 

\begin{table}[h!]
\caption{Score Network's training parameters. Note that in addition to the learning rate decay, the learning rate is decreased by factor of 0.1 every 30 epochs after the epoch number 120.}
\label{table:TrainingParamsScore}
\begin{tabular}{@{}c|cccc@{}}
\toprule
Parameter                   & Learning             & Learning       & Weight              & Momentum                 \\
 & Rate & Rate Decay & Decay & \\
\midrule
\multicolumn{1}{c|}{Values} & \multicolumn{1}{c|}{1e-3} & \multicolumn{1}{c|}{1e-7} & \multicolumn{1}{c|}{1e-5} & \multicolumn{1}{c|}{0.85} \\ \bottomrule
\end{tabular}
\end{table}
\subsubsection{Training Feature Detector}
Recently, deep convolutional neural networks have shown significant improvement over the state-of-the-art interest point detectors especially for detecting facial keypoints~\cite{sun2013deep}. In this paper, we propose to use a deep architecture for on-line keypoint detection. The proposed Network dubbed as "Feature Detector" uses the output of the left branch of the Feature Tracking Network to detect reliable features to track. Therefore, an additional fully connected network is attached to the output of the left branch of the network in order to classify each pixel as a keypoint or non-keypoint. Similar to the second stage of the training, during this stage of training, the parameters of the tracker are not updated as well.

Concerning training the network, we generated a train dataset by running roughly 100K points from KITTI Flow 2015 dataset~\cite{menze2015object} through the feature tracking network to obtain the ground truth labels for each pixel. To that end, pixels that were tracked correctly by the tracker are labeled as positive and otherwise negative and a balanced subset of these points are used to train the feature detection network. This ensures that the feature detector learns the behavior of the tracker on each point and can predict whether it will be reliably tracked or not. Such feature detection architecture can be used either to initialize the tracker or to re-initialize points in case the tracking is lost. The focus of this paper is mainly the feature tracker, thus, a comprehensive comparison of the proposed feature detector network against the state-of-the-art interest point detectors will be presented in another paper.

\section{Experimental Results}
This section presents extensive evaluation of the proposed unified feature detection and tracking framework. Different aspects of the proposed deep architecture is evaluated using challenging datasets such as KITTI FLOW 2015~\cite{menze2015object}, MIS dataset~\cite{puerto2014toward}, and UBC Patch dataset~\cite{UBCPatches}. The KITTI FLOW 2015 dataset is used to evaluate the tracking capabilities of the dataset under a real-world scenario for autonomous driving. The MIS dataset provides a more challenges mainly encountered in surgical vision such as large texture-less areas, specular highlights, large deformations, close distance to the scene, motion blur, blood, and smoke~\cite{parchami2014endoscopic}. On the other hand, the UBC patches dataset is employed to evaluate the feature tracker under a different application where the tracker is used to perform feature matching. Deep-PT is mainly compared against a modified version of the KLT-Tracker which is a widely used method for tracking in computer vision applications such as~\cite{ji2016robust},~\cite{singha2016accurate}, and~\cite{lim2017real}.

\subsection{Evaluation on KITTI Flow 2015}
The performance of the tracker is evaluated using KITTI Flow 2015 dataset over 30K points obtained by the following protocol. The KITTI dataset provides a semi-dense ground truth flow information for each pair of consecutive images. This ground truth data is used to generate roughly 30K pairs of corresponding points extracted around Harris corners and SIFT interest points in consecutive image pairs. Concerning comparison metrics, the tracking is compared by 1-pixel, 3-pixel, and 5-pixel accuracy where i-pixel accuracy means the ratio of correctly tracked pixels within "i" pixels of error over all pixel used for tracking.

\begin{figure}[th!]
\begin{center}
   \includegraphics[width=0.9\linewidth]{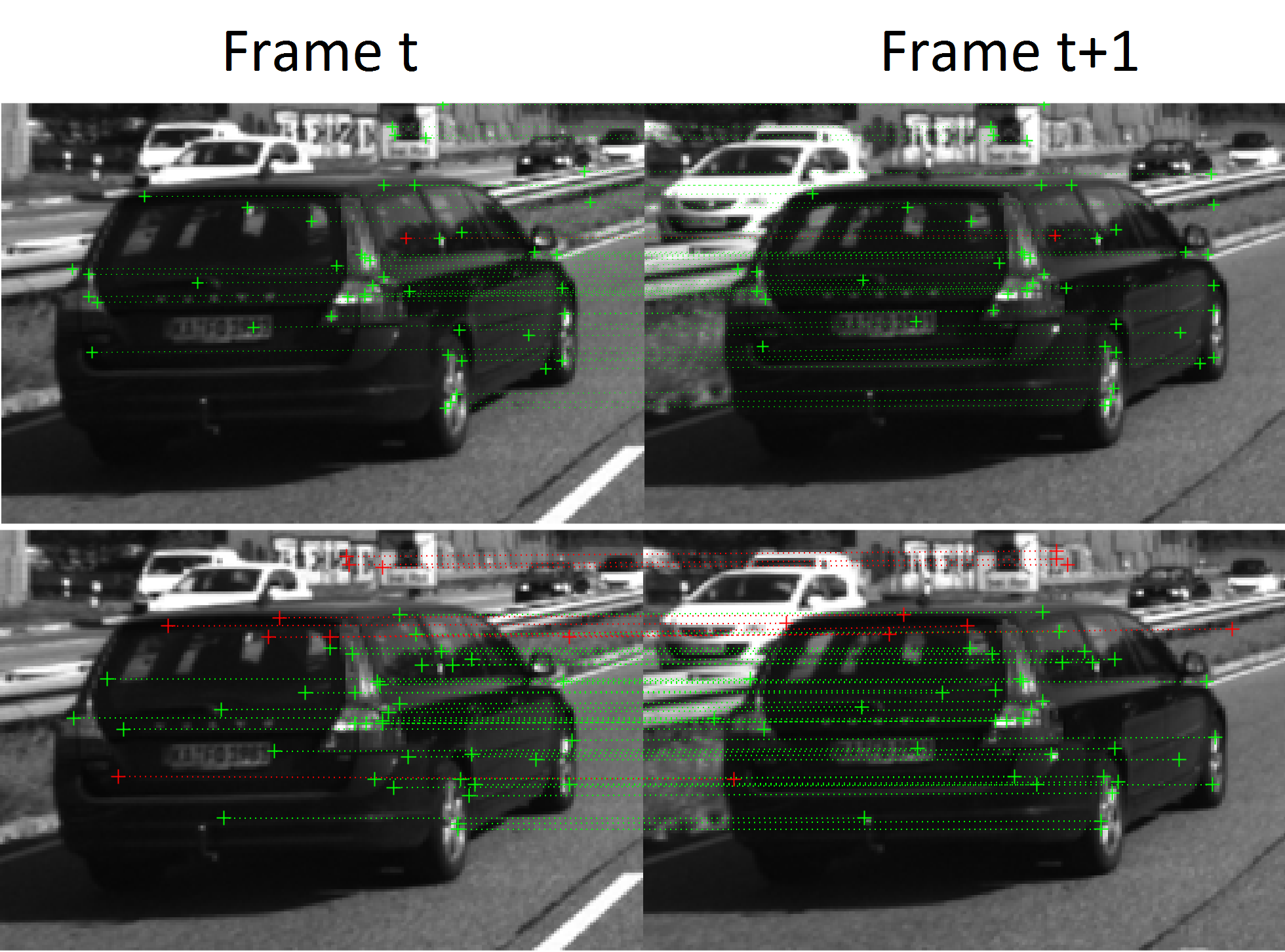}
\end{center}
   \caption{Qualitative comparison of Deep-PT Vs. forward-backward KLT-tracker where the lines show correspondences. Top row: visualization of the tracking performed by the Deep-PT over a cropped region of an image from KITTI Flow dataset. Bottom row: visualization of the tracking performed on the same image by the KLT-tracker}
\label{fig:TrackerVsKLT}
\end{figure}

The evaluation is performed by running the tracker specifically on these 30K points with the given ground truth and the results are compared against the most recent implementation of the KLT-Tracked algorithm with forward-backward error~\cite{kalal2010forward}. The forward-backward error ensures more reliable feature tracking by adopting a pyramidal approach for tracking both forward and backward in time. The points with high discrepancy in forward and backward tracking are marked as unreliable. Table.~\ref{tab:DeepVsKLT_KITTI} tabulates the accuracy of the proposed tracker compared against forward-backward error KLT-tracker. The results presented in Table.~\ref{tab:DeepVsKLT_KITTI} suggest a strong improvement over the state-of-the-art feature tracking methods.

\begin{table}[h!]
\centering
\caption{X-pixel tracking accuracy of Deep-PT and forward-backward KLT tracker in percentage.}
\label{table:DeepVsKLT_KITTI}
\begin{tabular}{@{}c|ccc@{}}
\toprule
Metric      & 1-pixel         & 2pixel       & 3-pixel \\
\midrule
\multicolumn{1}{c|}{Deep-PT} & \multicolumn{1}{c|}{$\%78.22$} & \multicolumn{1}{c|}{$\%88.78$} & \multicolumn{1}{c|}{$\%90.42$} \\ 
\midrule
\multicolumn{1}{c|}{KLT} & \multicolumn{1}{c|}{$\%53.93$} & \multicolumn{1}{c|}{$\%65.48$} & \multicolumn{1}{c|}{$\%70.61$} \\ \bottomrule
\end{tabular}
\end{table}

Fig.~\ref{fig:TrackerVsKLT} visualizes an example of tracking performed by our proposed method versus the KLT-tracker. In this figure, only a cropped region of the image is presented for convenience and green represents successful tracking of a point and red represents failure in tracking. The mis-tracked features detected by the tracking score network are not visualized here. As shown in Fig.~\ref{fig:TrackerVsKLT}, Deep-PT outperforms KLT tracker in effectively localizing features in the next frame. More specifically, the proposed method performs well on generic features and does not rely only on corner to predict the motion of a pixel. A closer look at Fig.~\ref{fig:TrackerVsKLT} reveals that the only mis-tracked point in the first row is actually tracked correctly in that local area considering the shadow on the car moves backwards.

\subsection{Evaluation on MIS dataset}
While the KITTI Flow 2015 dataset provides a great ground truth data for our tracking purpose, it has limited types of motion and challenges. Thus, we propose to perform an experiment under a Minimally Invasive Surgical environment where the images are captured using an endoscopic camera of the da Vinci surgical platform~\cite{puerto2014toward}. Such dataset imposes more challenges, however, it lacks ground truth data fro tracking.

\begin{table}[h!]
\centering
\caption{Pixel back-projection error and inlier percentage for the MIS dataset.}
\label{table:BackProj}
\begin{tabular}{@{}c|cc@{}}
\toprule
      & Average Error         & \% Inlier \\
\midrule
\multicolumn{1}{c|}{Lowe's} & \multicolumn{1}{c|}{$4.66\pm4.24$} & \multicolumn{1}{c|}{$\%34$} \\
\midrule
\multicolumn{1}{c|}{AMA} & \multicolumn{1}{c|}{$2.49\pm2.42$} & \multicolumn{1}{c|}{$\%40$} \\
\midrule
\multicolumn{1}{c|}{Cho} & \multicolumn{1}{c|}{$3.56\pm3.35$} & \multicolumn{1}{c|}{$\%39$} \\
\midrule
\multicolumn{1}{c|}{HMA} & \multicolumn{1}{c|}{$2.84\pm2.64$} & \multicolumn{1}{c|}{$\%39$} \\
\midrule
\multicolumn{1}{c|}{Deep-PT} & \multicolumn{1}{c|}{$2.71\pm2.81$}  & \multicolumn{1}{c|}{$\%82$}\\ \bottomrule
\end{tabular}
\end{table}

The quantitative evaluation the MIS dataset is performed by following the same protocol provided by~\cite{puerto2014toward}. For this purpose, the methods are compared using a back-projection error metric where the points in the current frame are back-projected to the previous frame using homography and the euclidean distance between the corresponding points is considered as error measure. Homography matrices are computed by considering the same planar patches obtained by~\cite{puerto2012hierarchical}. Table~\ref{table:BackProj} presents the back-projection error for the proposed method, Hierarchical Multi-Affine (HMA)~\cite{puerto2012hierarchical} feature matching, Lowe's~\cite{lowe2004distinctive}, Adaptive Multi-Affine (AMA)~\cite{souza2011adaptive} and Cho~\cite{cho2009feature}. As Table ~\ref{table:BackProj} suggests, Deep-PT provides more inlier points with a higher accuracy than the state-of-the-art methods in surgical environment. 

\begin{figure}[th!]
\begin{center}
   \includegraphics[width=0.9\linewidth]{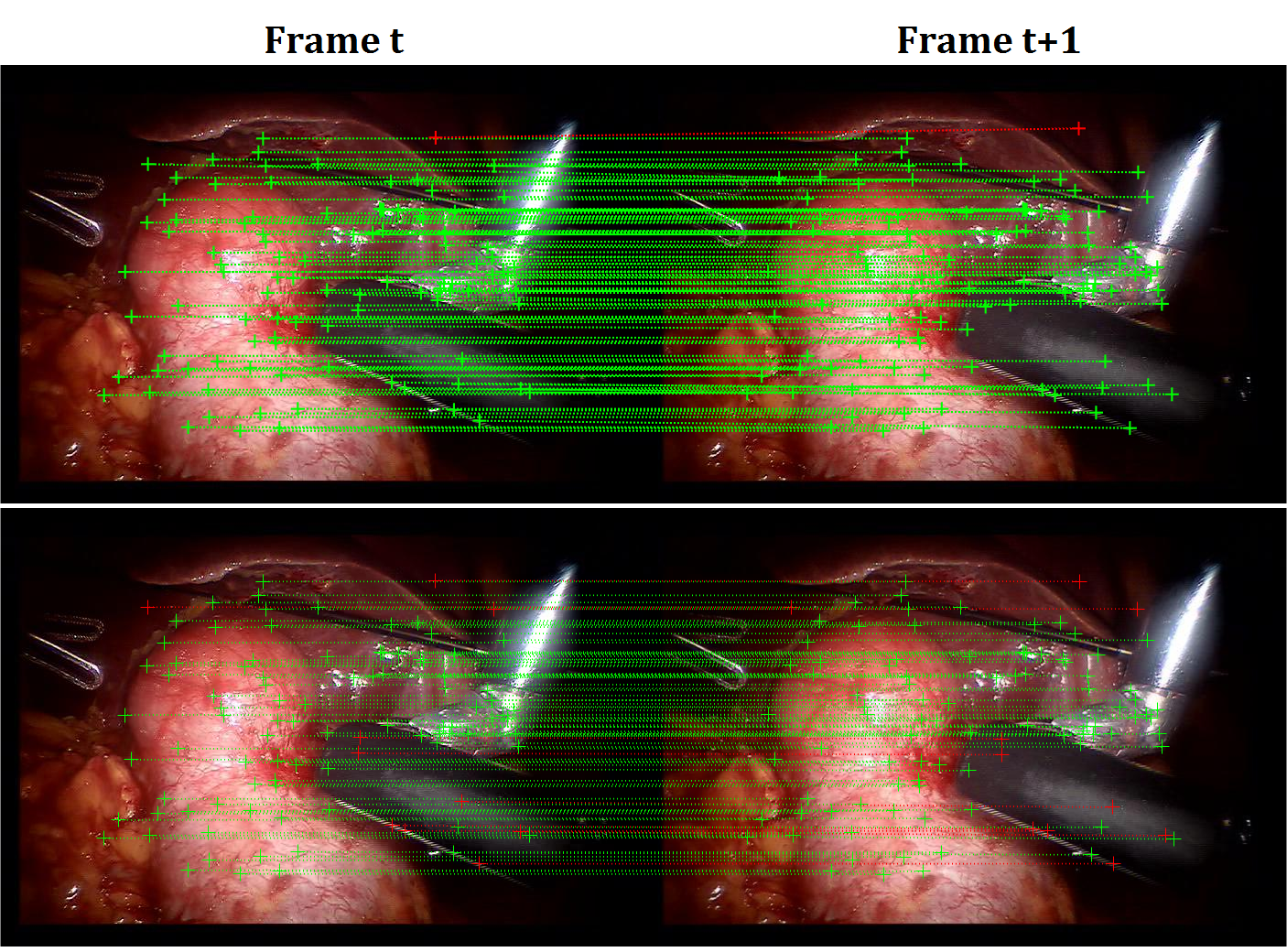}
\end{center}
   \caption{Qualitative comparison of Deep-PT Vs. forward-backward KLT-tracker where the lines show correspondences. Top row: visualization of the tracking performed by the Deep-PT over a pair of consecutive frames from the MIS dataset. Bottom row: visualization of the tracking performed on the same images by the KLT-tracker}
\label{fig:SurgicalRes}
\end{figure}

Fig.~\ref{fig:SurgicalRes} presents a pair of images from the MIS dataset where the feature points are visualized on each image. In Fig.~\ref{fig:SurgicalRes}, the correctly tracked features are visualized in green whereas the mis-tracked features that were not detected by the tracking score network are visualized in red. As suggested by Fig.~\ref{fig:SurgicalRes}, the proposed method performs better in such texture-less environments than the KLT-tracker.

\subsection{Evaluation on UBC Patches dataset}
So far the tracking capabilities of the proposed Deep-PT is evaluated and in this section we tend to evaluate the patch-matching competence of the proposed method against the state-of-the-art deep learning based methods. To that end, the UBC Patches dataset is employed to compare small patches. The trained feature score matching network is responsible to generate a matching score between two given patches. 

In order to compare different algorithms fairly, we followed the protocol suggested by~\cite{han2015matchnet} and the error rate at $\%95$ recall is reported in percentage. Table.~\ref{table:DeepVsKLT_UBC} tabulates the comparison of the proposed method against MatchNet and other recent local descriptor learning algorithms. Considering that Deep-PT is not trained specifically to classify patches to matching and non-matching categories, the performance of the network is satisfactory. Additionally, Deep-PT utilizes only a small patch inside the $64\times64$ patches from the dataset and training the network with the whole patches would noticeably increase the accuracy of matching.

\begin{table}[h!]
\centering
\caption{UBC matching results. Numbers are Error at \%95 recall in percentage.}
\label{table:DeepVsKLT_UBC}
\begin{tabular}{@{}c|cc@{}}
\toprule
Training      & Notredame         & Liberty \\
\cmidrule(lr){2-3}
Training      & Liberty         & Notredame \\
\midrule
\multicolumn{1}{c|}{Baseline: nSift+NNet~\cite{han2015matchnet}} & \multicolumn{1}{c|}{$\%20.44$} & \multicolumn{1}{c|}{$\%14.35$} \\
\midrule
\multicolumn{1}{c|}{Trzcinski et al~\cite{trzcinski2012learning}} & \multicolumn{1}{c|}{$\%18.05$} & \multicolumn{1}{c|}{$\%14.15$}\\ 
\midrule
\multicolumn{1}{c|}{Brown et al~\cite{brown2011discriminative}} & \multicolumn{1}{c|}{$\%16.85$} & \multicolumn{1}{c|}{$N.A.$}\\ 
\midrule
\multicolumn{1}{c|}{Simonyan et al~\cite{simonyan2014learning}} & \multicolumn{1}{c|}{$\%16.56$} & \multicolumn{1}{c|}{$\%9.88$}\\ 
\midrule
\multicolumn{1}{c|}{MatchNet~\cite{han2015matchnet}} & \multicolumn{1}{c|}{$\%9.82$} & \multicolumn{1}{c|}{$\%5.02$}\\ 
\midrule
\multicolumn{1}{c|}{Deep-PT} & \multicolumn{1}{c|}{$\%15.99$} & \multicolumn{1}{c|}{$\%12.79$}\\ 
\bottomrule
\end{tabular}
\end{table}

\section{Conclusion}
This paper presented a novel unified deep learning based pixel tracking  framework capable of detecting good features to track and re-initialize new features in case of failure in tracking. In that regard, Deep-PT intuitively simulates cross-correlation in deep learning to localize a pixel in the next time frame. The ability to detect features that are more suitable for the trained tracker differentiates the proposed methods from the state-of-the-art methods. Moreover, the results on KITTI Flow 2015 and MIS dataset suggests that in a real-world scenario, Deep-PT outperforms existing methods and can be generalized to any type of environment such as outdoors and surgical images. Additionally, extensive comparisons on UBC Patch dataset against patch-matching algorithms suggests that the network can be generalized to similar problems. Deep-PT is a reliable method for tracking features based on a learning method which enables it to track a variety of reliable types of features more accurately.

The proposed method is not perfect and has defects. More specifically Deep-PT fails in environments with highly repetitive texture patterns as suggested by experiments. The next step is to train the feature detection network to avoid such pitfalls. Moreover, a more extensive comparison of the feature detector with the state-of-the-art interest point detection algorithms will be performed. Additionally, a study on long-term tracking capabilities of the system should be explored and addressed in the future studies.

{\small
\bibliographystyle{ieee}
\bibliography{egbib}
}

\end{document}